\documentclass[letterpaper, 10 pt, conference]{ieeeconf}  

\IEEEoverridecommandlockouts                             
\usepackage{graphics} 
\usepackage{epsfig} 
\usepackage{mathptmx} 
\usepackage{times} 
\usepackage{amsmath} 
\usepackage{amssymb}  
\usepackage[linesnumbered,ruled,vlined]{algorithm2e}
\usepackage{multirow}
\usepackage{multicol}
\usepackage{makecell}
\usepackage{subcaption}
\usepackage{hyperref}

\usepackage{geometry}
\title{\LARGE \bf
SemanticFormer: Holistic and Semantic Traffic Scene Representation for Trajectory Prediction using Knowledge Graphs
}

\author{Zhigang Sun$^{1}$, Zixu Wang$^{2,3}$, Lavdim Halilaj$^{3}$, Juergen Luettin$^{3}$
\thanks{$^{1}$Zhigang Sun is with Bosch Center for Artificial Intelligence, (Corresponding author: Zhigang Sun)
        {\tt\small zhigang.sun3@cn.bosch.com}, {\tt\small zhigang.sun20@alumni.imperial.ac.uk} }
\thanks{$^{3}$Zixu Wang, Lavdim Halilaj, Juergen Luettin are with Robert Bosch GmbH
        {\tt\small \{firstname.lastname\}@bosch.com}}%
\thanks{$^{2}$Zixu Wang is with the Technical University of Munich (TUM), Germany
        {\tt\small zixu.wang@tum.de}}
}

\begin{document}
\maketitle
\begin{abstract}

Trajectory prediction in autonomous driving relies on accurate representation of all relevant contexts of the driving scene, including traffic participants, road topology, traffic signs, as well as their semantic relations to each other. 
Despite increased attention to this issue, most approaches in trajectory prediction do not consider all of these factors sufficiently. 
We present \textbf{SemanticFormer}, an approach for predicting multimodal trajectories by reasoning over a semantic traffic scene graph using a hybrid approach. 
It utilizes high-level information in the form of meta-paths, i.e. trajectories on which an agent is allowed to drive from a knowledge graph which is then processed by a novel pipeline based on multiple attention mechanisms to predict accurate trajectories. 
SemanticFormer comprises a hierarchical heterogeneous graph encoder to capture spatio-temporal and relational information across agents as well as between agents and road elements.
Further, it includes a predictor to fuse different encodings and decode trajectories with probabilities. 
Finally, a refinement module assesses permitted meta-paths of trajectories and speed profiles to obtain final predicted trajectories. Evaluation of the nuScenes benchmark demonstrates improved performance compared to several SOTA methods. 
In addition, we demonstrate that our knowledge graph can be easily added to two graph-based existing SOTA methods, namely VectorNet and LaFormer, replacing their original homogeneous graphs. 
The evaluation results suggest that by adding our knowledge graph the performance of the original methods is enhanced by 5\% and 4\%, respectively. Graph data is available at  \url{https://github.com/boschresearch/nuScenes_Knowledge_Graph}

\end{abstract}

\section{INTRODUCTION}

Autonomous vehicles are recognized as a promising solution to address critical challenges such as road safety, traffic congestion, and energy optimization.
A crucial task towards the realization of autonomous driving vision is motion prediction\cite{9304764}. 
It involves determining a set of spatial coordinates that represent the predicted movement of a given agent within a future time window.
However, motion prediction is a challenging task due to various contextual factors such as the difficulty of intention prediction, the complex interactions of traffic participants, the intricate road topology, comprising lanes, lane dividers, and pedestrian crossings, as well as adherence to traffic regulations. 
State-of-the-art approaches utilize various representations for traffic scenes such as raster-based~\cite{Cui2018MultimodalTP,PhanMinh2019CoverNetMB}, or graph-based~\cite{Li2020EvolveGraphMT,Gao2020VectorNetEH} to capture and utilize contextual information sufficiently. 

\begin{figure}[t]
    \centering
    \includegraphics
    [width=0.49\textwidth]
    {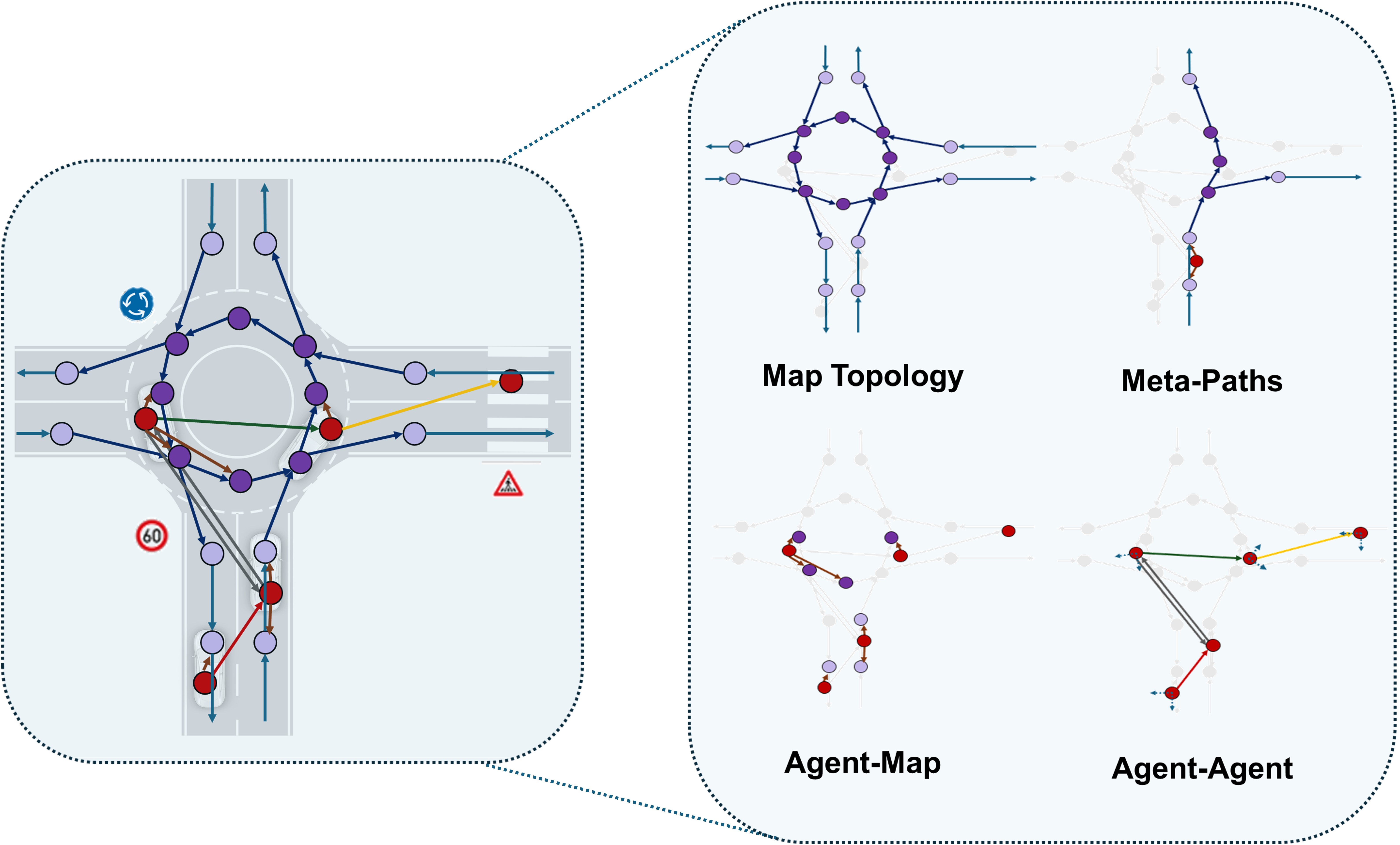} 
    \caption{
    Driving scenes represented in a heterogeneous graph capturing all relevant map details, traffic agents, and their semantic relationships.
    }
    \vspace{-7.5pt}
    \label{Overview}
\end{figure}

Recent work applies a knowledge graph (KG) to encode diverse contextual information from traffic scenes~\cite{iccvw_leon}. 
Figure~\ref{Overview} illustrates various types of elements comprised in a typical traffic scene including different entities and their relations along with their semantic descriptions.
We propose a novel approach that leverages heterogeneous information of static and dynamic elements modeled in the KG. 
It contains an attention mechanism for consuming semantic relationships and dependencies between traffic agents and road elements for accurate multimodal trajectory prediction.
Main contributions:

\begin{itemize}
     \item 
     A knowledge graph based approach to encode all relevant static and dynamic elements of a traffic scene with their semantic relationships.
    \item 
    A hybrid architecture with attention mechanisms to model the semantic relationships and dependencies between traffic agents and road elements for accurate multi-modal trajectory prediction. Evaluated on nuScenes dataset~\cite{Caesar2020nuscenes}.
    \item Dedicated experiments to demonstrate the easiness of incorporating our KG into existing graph-based trajectory prediction models. Concretely, we integrate the KG into VectorNet~\cite{Gao2020VectorNetEH} and LaFormer~\cite{Liu2023LAformerTP} (changing GIG block). The evaluation results show that incorporating KG with VectorNet and LaFormer helps improve their ADE performance by 5\% and 4\%, respectively.

\end{itemize}

\begin{figure*}[t]
    \centering
\includegraphics[width=\textwidth]{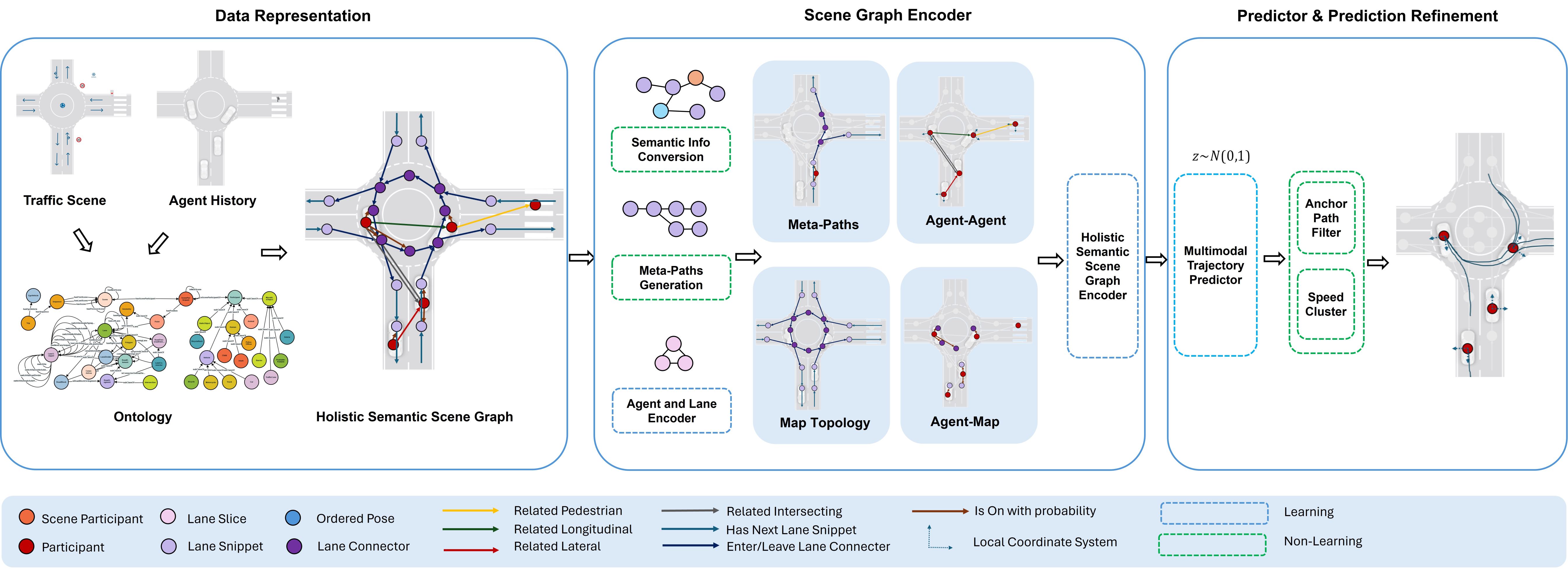} 
    \caption{SemanticFormer Overview: \textit{Data Representation} models the static map information and dynamic agents interaction by a holistic \textit{knowledge graph}. \textit{Scene Graph Encoder} extracts \textit{meta-paths} and generates holistic \textit{latent representation} for agents and lanes. \textit{Probability Predictor} fuses the encodings and outputs trajectory candidates. \textit{Prediction Refinement} uses \textit{anchor paths} and \textit{speed profiles} to evaluate trajectories and generates final predictions.}
    \vspace{-10pt}
    \label{fig:example}
\end{figure*}

\section{RELATED WORK}

\textbf{Representation}. 
Early methods for trajectory prediction use raster-based birds-eye-view representations of the map and agents encoding them with a number of channels for different information sources~\cite{Djuric2018UncertaintyawareSM,Hong2019RulesOT}.
These methods are extended to predict multiple trajectories with associated  probabilities\cite{Cui2018MultimodalTP,PhanMinh2019CoverNetMB}.
Others aim to estimate probability distribution heat maps representing locations where agents could be located at a fixed time horizon~\cite{Gilles2021HOME,Gilles2021THOMASTH}.
However, these models usually do not have access to high-level information and need to learn complex relationships from raw pixels.

Graph-based approaches represent scenes as vectors, polylines and graphs and thus operate at a higher level of abstraction~\cite{Gao2020VectorNetEH,Liang2020Learning,Casas2019SpAGNNSG, Varadarajan2021MultiPathEI,Konev2022MPAMB,Liu2023LAformerTP}. 
VectorNet \cite{Gao2020VectorNetEH} encodes both map features and agent trajectories as polylines and then merges them with a global interaction graph.  
TNT~\cite{Zhao2020TNTTT} extends VectorNet and combines it with multiple target reference trajectory proposals sampled from the lanes to diversify the prediction points. 
Unfortunately, these techniques usually use homogeneous graphs with one entity type and one relation type which prevents them from representing the rich heterogeneous traffic scene along with their complex relations. 

Methods that use heterogeneous graphs, i.e. graphs with different entity types such as vehicles, bicycles or pedestrians and relation types like agent-to-lane or agent-to-agent, are recently proposed~\cite{Mo2022MultiAgentTP,Jia2022HDGTHD,Monninger2023SCENE,Wonsak2022Multi, Grimm2023HeterogeneousGT, wang2024socialformersocialinteractionmodeling}.
However, they are limited to only a portion of the relevant information and are unable to fully capture all scene details and the interconnections between the entities.
Our approach aims to fill this gap using formal ontologies for constructing a knowledge graph~\cite{Halilaj2022KnowledgeGF,Halilaj2023KnowledgeGI,Luettin2022Survey, xiong2024tilp, xiong2024large} capturing the rich information of traffic scenes. 
Knowledge graphs have been applied in other automotive applications like POI recommendation~\cite{Halilaj2021TowardsAK,Werner2020RETRART} and driving situation understanding~\cite{Halilaj2021AKG}.   

\textbf{Encoding}. 
Early encodings are based on {CNNs}~\cite{Chai2019MultiPathMP,Casas2018IntentNetLT,Cui2018MultimodalTP,Hong2019RulesOT}, while more recent works use {GNN}s~\cite{Casas2019SpAGNNSG,Gao2020VectorNetEH,Liang2020Learning,Konev2022MPAMB,Varadarajan2021MultiPathEI}.
Attention mechanisms have recently attracted high interest in modeling the interactive behavior between agents for raster-based 
approaches~\cite{Tang2019MultipleFP,Messaoud2020Attention,Park2020DiverseAA,Yuan2021AgentFormerAT,Girgis2021LatentVS}, graph-based approaches~\cite{Khandelwal2022What,Liu2021Multimodal,Huang2021MultimodalMP,Ngiam2022SceneTA,Nayakanti2022WayformerMF,Shi2022MotionTW} and map-free approaches~\cite{Mercat2019MultiHeadAF}.
A hierarchical vector transformer-based approach, {HiVTHV} is presented in~\cite{Zhou2022HiVTHV} that consists of a local context feature encoding followed by the global message passing among agent-centric local regions. 
Autoregressive trajectory prediction approaches generating trajectories at intervals to produce scene-consistent multi-agent trajectories are proposed in~\cite{Rhinehart2019PRECOGPC,Tang2019MultipleFP,Amirloo2022LatentFormerMT,salzmann2020trajectron++,Yuan2021AgentFormerAT}.
Based on language modeling concepts with transformers, Motion{LM}~\cite{Seff2023MotionLMMM} treats continuous trajectories as sequences of discrete motion tokens and cast multi-agent motion prediction as a language modeling task.
In~\cite{Keysan2023CanYT}, a pretrained language model is used to encode text describing traffic situations combined with raster-based encodings.
A game-theoretic modeling and learning approach considering relations between scene elements, alongside a novel hierarchical transformer decoder architecture is presented in~\cite{Huang2023GameFormerGM}.
We also use a transformer-based architecture but encode different information sources including map topology, meta-paths, as well as relational information.  

\begin{figure*}[h]
    \centering
\includegraphics[width=\textwidth]{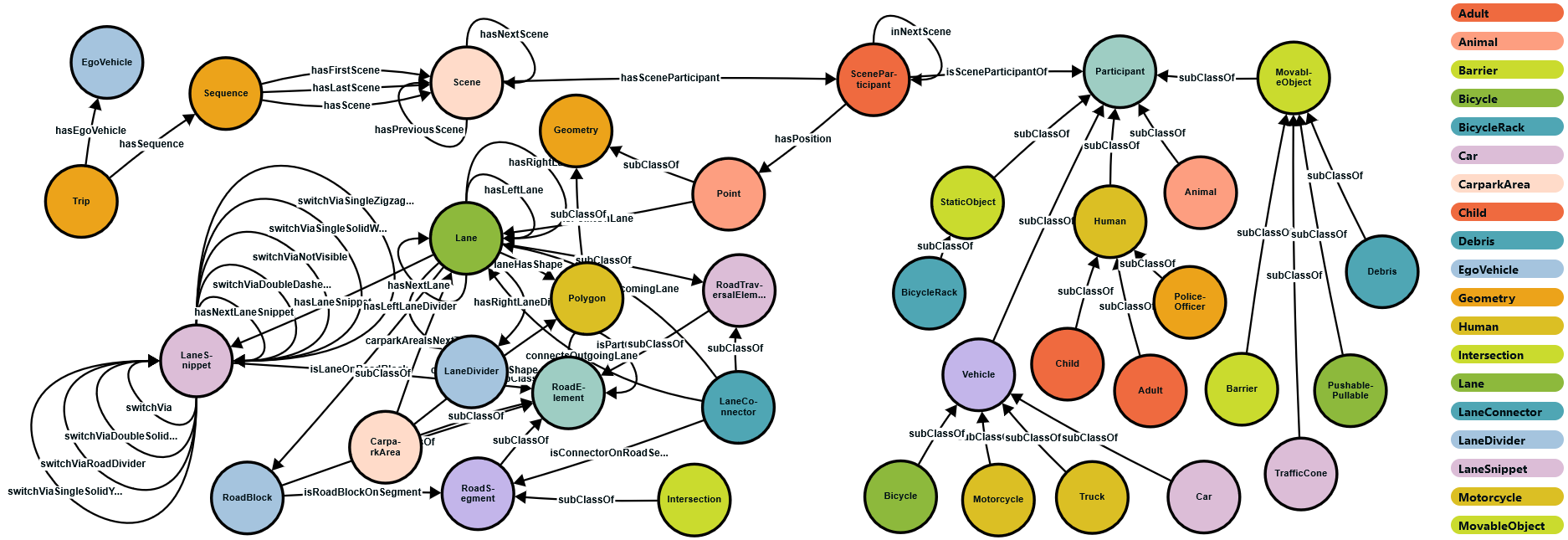} 
  \caption{Illustration of traffic scene ontologies~\cite{iccvw_leon}: \textit{Agent Ontology} defines agent attributes like \textit{category, speed, position, and trajectory}, and relationships to map like \textit{distance to lane, and path distance}. \textit{Map Ontology} defines map elements like \textit{lane snippet, lane slice, traffic light, etc.}, and relations within map elements like \textit{left/right lane, switch via double dashed line}.}
  \vspace{-10pt}
  \label{fig:ontology}
\end{figure*}

\textbf{Predicting}. 
Goal- or intention conditioned systems sample goal candidates and predict trajectories conditioned on them~\cite{Casas2020ImplicitLV,Albrecht2020InterpretableGP,Shi2022MotionTW,Zhao2020TNTTT,Gu2021DenseTNTET}.
Grid-based policy learning via maximum entropy inverse reinforcement learning is used in~\cite{Deo2020TrajectoryFI} to condition trajectory forecasts.
Authors in ~\cite{Lu2022KEMPKH} use key-frames as representative states to trace out the general direction of the trajectory.
Approaches considering lane-aware scene constraints that align motion dynamics with scene information are shown in~\cite{Liu2023LAformerTP,Park2023LeveragingFR}.
Our architecture is related, but we use a heterogeneous graph transformer to process the heterogeneous information of the KG.
Others use anchors, fixed sets of anchor trajectories corresponding to permitted trajectories, to guide trajectory prediction~\cite{Chai2019MultiPathMP,PhanMinh2019CoverNetMB,Grimm2023HeterogeneousGT,naumann2023lanelet2}.
\cite{Varadarajan2021MultiPathEI} presents a method to learn latent representations of anchor trajectories.
Query-centric trajectory prediction is proposed in~\cite{Zhou2023QueryCentricTP,chen2023q}, where agents' decisions are formulated as information queries using the available information before they make a decision. 
Our approach is related but refines anchors into meta-paths by using contextual information.

\section{METHODOLOGY}
We aim to represent all relevant information that governs the behavior of traffic participants. 
For example, information about lane dividers (e.g. dashed line, solid line), conveys information about permitted lane changes and is therefore important for trajectory prediction; a pedestrian crossing together with the distance and direction of nearby pedestrians governs the behavior of oncoming vehicles. 
As seen below, it is not only important to represent all relevant information but also their relational information.
We address this challenge by representing the map and agents with a knowledge graph. 
This enables us to explicitly model the various map elements and agents as well as their semantic relations.  
It also allows for the modeling of diverse traffic agents types like cars, and bicycles, and their relations in driving situations such as whether two agents might interact, or drive behind or next to one another.

In the following, we describe a comprehensive architecture depicted in Figure~\ref{fig:example}, which uses a knowledge graph for predicting multimodal trajectories. 
The architecture begins by taking the scene graph $g_i$ as input and outputs multimodal trajectories for the target agent. 
Finally, the refinement module filters the predicted trajectories, considering anchor paths and speed profiles to avoid failure cases.
\vspace{-5pt}
\subsection{Ontology and Heterogeneous Scene Graph} 
\vspace{-5pt}
We utilize ontologies to explicitly represent the abundance of information from traffic scenes. 
Thus, based on the domain knowledge we model relationships between entities considered important for the trajectory prediction task. 
Figure~\ref{fig:ontology} illustrates the developed ontologies, encompassing various entity and relation types. 
The entity types are categorized into two groups: the first one contains static entities like lane types, boundaries, center lines and stop areas; the second group contains dynamic entities like agents, their states, and bounding boxes. 
As for relation types, they fall into three groups: 1) between agents, which construct semantic associations such as lateral, longitudinal, and intersecting, as shown in Figure~\ref{fig:agent-agent} akin to the concepts presented in~\cite{Zipfl2022RelationbasedMP}; 2) between map elements, establishing lane connectivity and relationships between lanes and road infrastructure elements like stop areas, traffic lights, pedestrian crossings; and 3) relations between map elements and agents, utilizing probability projection to map agents onto road infrastructure.
Based on the designed ontology, we represent the scene by a heterogeneous scene graph $G=(V, E, \tau, \phi)$. 
It has nodes $v \in V$, their types $\tau(v)$, and edges $(u, v) \in E$, with edge types $\phi(u, v)$. The edges are directed since they are based on properties of the knowledge graph.  
\vspace{-5pt}
\subsection{Problem Formulation for Trajectory Prediction}
\vspace{-5pt}
We assume that the perception part can provide detailed information about agent positions, and past motion as well as the HD map, so we build the scene graph as described in the previous section. 
Then, a sample of the dataset can be formed as $\left(g_i, y_i\right)$  where $g_i$ is a sample scene graph with trajectory information, local map, and target identifier and $y_i$ is the ground truth future trajectory of the given target.
Both agent past trajectories and map information are represented hierarchically. 
Further, $g_i \in G$ covers the information within a chosen time horizon $\left\{-t_h+1, \cdots, 0,1, \cdots, t_f\right\}$. 
We use $\mathbf{P}_{-t_h+1: 0}^i=\left\{sp_{-t_h+2}^i, sp_{-t_h+3}^i, \ldots, sp_0^i\right\}$ to represent respective scene participant nodes. 
Each participant node $sp_t^i$ is modeled as $sp_t^i=\left[d_{t, s}^i, d_{t, e}^i, a^i\right]$, where $d_{t, s}^i$ and $d_{t, e}^i$ stands for previous and current time stamps participant locations, whereas $a^i$ represent additional attributes like velocity, acceleration, heading change rate and the object type. 
For map information we use $\mathbf{S}_{1: N}^i=\left\{s_1^i, s_2^i, \ldots, s_N^i\right\}$ to denote a lane snippet, $s_n^i$ for lane slices and $N$ the length of the given lane snippet. 
Each lane slice vector $s_n^i=\left[d_{n, s}^i, d_{n, e}^i, a_i, d_{n, \mathrm{pre}}^i\right]$ adds $d_{n, \mathrm{pre}}^i$ to indicate the predecessor of the starting point. 
Connections between lane snippets are built by lane connectors $\mathbf{C}_{1: N}^i=\left\{c_1^i, c_2^i, \ldots, c_N^i\right\}$, where each $c_n^i$ encodes an ordered pose inside the lane connector of length $N$.

Coordinates in the knowledge graph are initially in a global coordinate system.
These are then transformed separately into local, scene graph-specific coordinates, with the origin at the location of the target agent and the positive y-axis pointing along the facing direction of the target. 

\subsection{Semantic Scene Graph Hierarchical Modeling}
\subsubsection{Meta-Path Generation}
We extract meta-paths to describe permitted and possible driving directions to navigate the target participant. 
Meta-paths related to the permitted lane changes and turns can be divided into three groups: 1) lane-changing; 2) entering the lane connector; and 3) leaving the lane connector.
\begin{figure}[t]
  \centering
  \begin{subfigure}[b]{0.23\textwidth}
    \includegraphics[width=\textwidth]{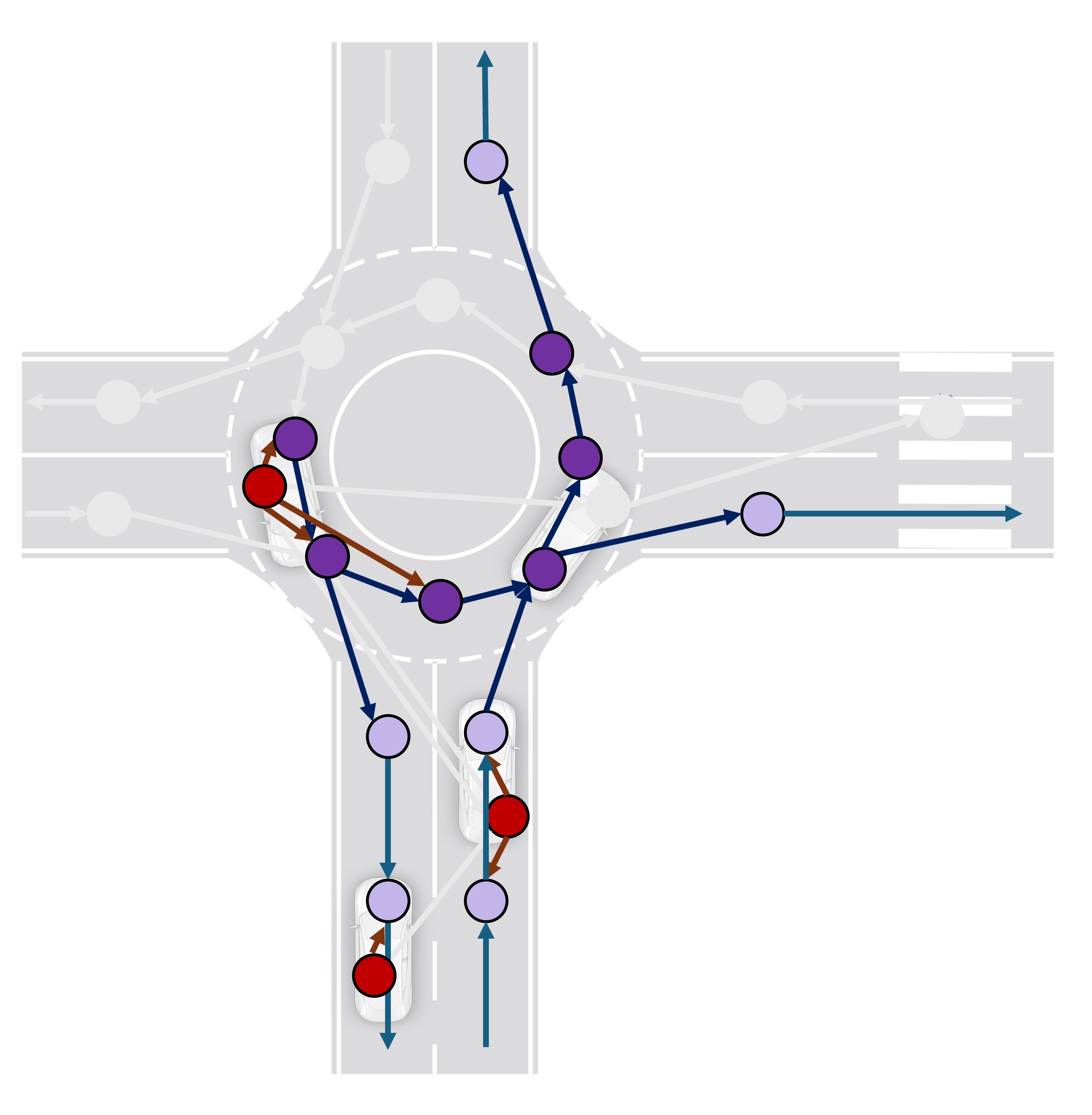}
    \caption{Meta-path Generation}
    \label{fig:meta_path}
  \end{subfigure}
  \hfill
  \begin{subfigure}[b]{0.23\textwidth}
    \includegraphics[width=\textwidth]{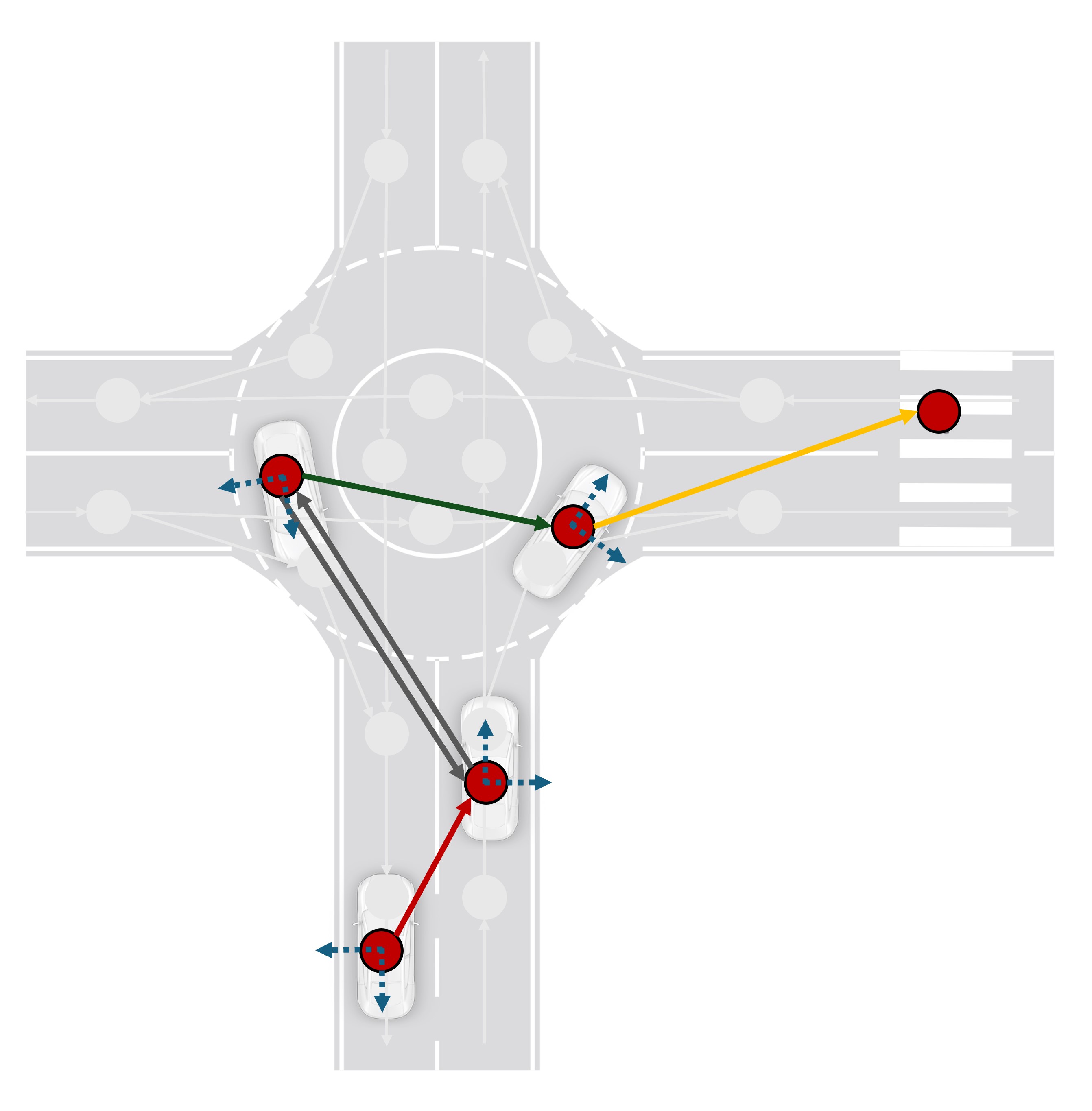}
    \caption{Agent-Agent Interaction}
    \label{fig:agent-agent}
  \end{subfigure}
  \caption{(a) Illustration of meta-paths depicting permitted trajectories. (b) Illustration of the participant interaction graph: Characterized by edge types: \textit{Longitudinal(green)}, \textit{Intersecting(gray)}, \textit{Lateral(red)}, and \textit{Pedestrian(yellow)}.}
    \vspace{-10pt}
\end{figure}
Figure~\ref{fig:meta_path} gives a qualitative analysis of generated meta-paths.
Specifically, we illustrate sample meta-paths below, such as lane-changing~\ref{eq:1}, leaving connector~\ref{eq:2}, and entering connector cases~\ref{eq:3}, where $\Phi$ represents the meta-path.
\begin{equation} 
\Phi_0 = \mathbf{P} \stackrel{isOn}{\longrightarrow} \mathbf{S} \stackrel{switchViaX}{\longrightarrow} \mathbf{S} \stackrel{switchViaX}{\longrightarrow} \mathbf{S} 
\label{eq:1}
\end{equation}
\vspace{-14pt}
\begin{equation}
 \Phi_1 = \mathbf{P} \stackrel{isOn}{\longrightarrow} \mathbf{C} \stackrel{CconnectS}{\longrightarrow} \mathbf{S} \stackrel{switchViaX}{\longrightarrow} \mathbf{S} 
 \label{eq:2}
\end{equation}
\vspace{-14pt}
\begin{equation}
\Phi_2 = \mathbf{P} \stackrel{isOn}{\longrightarrow} \mathbf{S} \stackrel{switchViaX}{\longrightarrow} \mathbf{S} \stackrel{SconnectC}{\longrightarrow} \mathbf{C} 
\label{eq:3}
\end{equation}
\vspace{-14pt}
\subsubsection{Agent Motion and Lane Encoder}
\begin{figure}[t]
  \centering
  \includegraphics[width=0.45\textwidth]{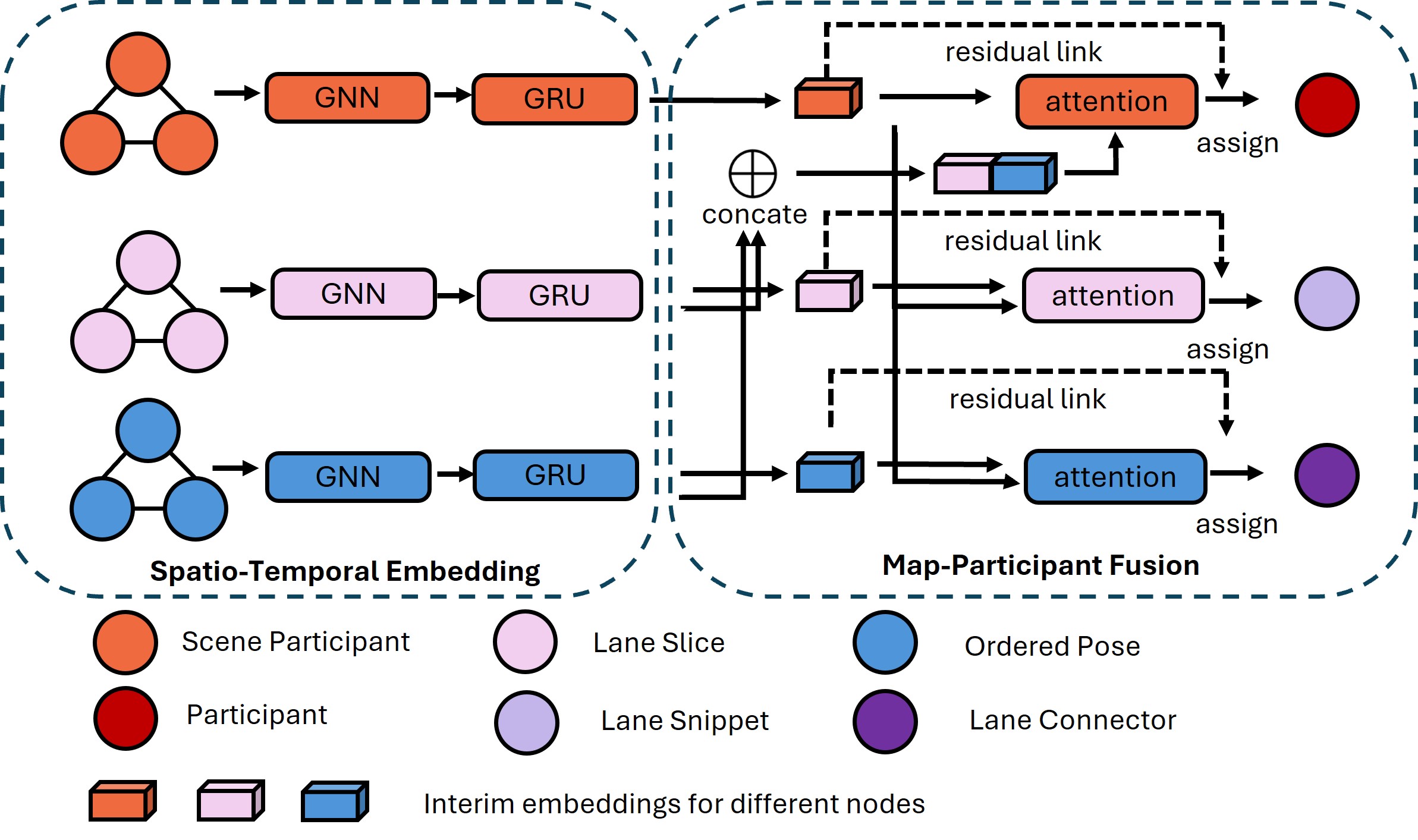} 
  \caption{Illustration of the agent motion and lane encoder: GNN and GRU extracts spatio-temporal information, attention mechanism models participants related lane.}
    \vspace{-10pt}
  \label{fig:agent-map-encoder}
\end{figure}

This component is responsible for encoding spatio-temporal information. 
We process participants $\mathbf{P}^i$, lane snippets $\mathbf{S}_{1: N}^i$, and lane connectors $\mathbf{C}_{1: N}^i$ in a sequential manner using both a Graph Neural Network (GNN) and a Gated Recurrent Unit (GRU) layer. 
Their respective encodings are represented by $p_i$, $s_j$, and $c_z$. 
Further, inspired by LaneGCN~\cite{Liang2020Learning}, we merge the outcomes as shown in Figure~\ref{fig:agent-map-encoder}. 
Equation~\ref{eq:4} introduces lane information to the related agents while equation~\ref{eq:5} and equation~\ref{eq:6} add participant information to the related lanes and lane connectors. 
\begin{equation}
p_i=p_i+\operatorname{CrossAtt}\left\{p_i, [s_j, c_z]\right\}
\label{eq:4}
\end{equation}
\vspace{-14pt}
\begin{equation}
s_j=s_j+\operatorname{CrossAtt}\left\{s_j, p_i\right\}
\label{eq:5}
\end{equation}
\vspace{-14pt}
\begin{equation}
c_z=c_z+\operatorname{CrossAtt}\left\{c_z, p_i\right\}
\label{eq:6}
\end{equation}
where $i \in\left\{1, \ldots, N_{\text {P}}\right\}, j \in\left\{1, \ldots, N_{\text {LS }}\right\}, z \in\left\{1, \ldots, N_{\text {LC }}\right\}$. 
Encodings are assigned to node attributes in scene graph $g_i$.

\subsubsection{Scene Graph Encoder}
\begin{figure}[t]
  \centering
  \includegraphics[width=0.45\textwidth]{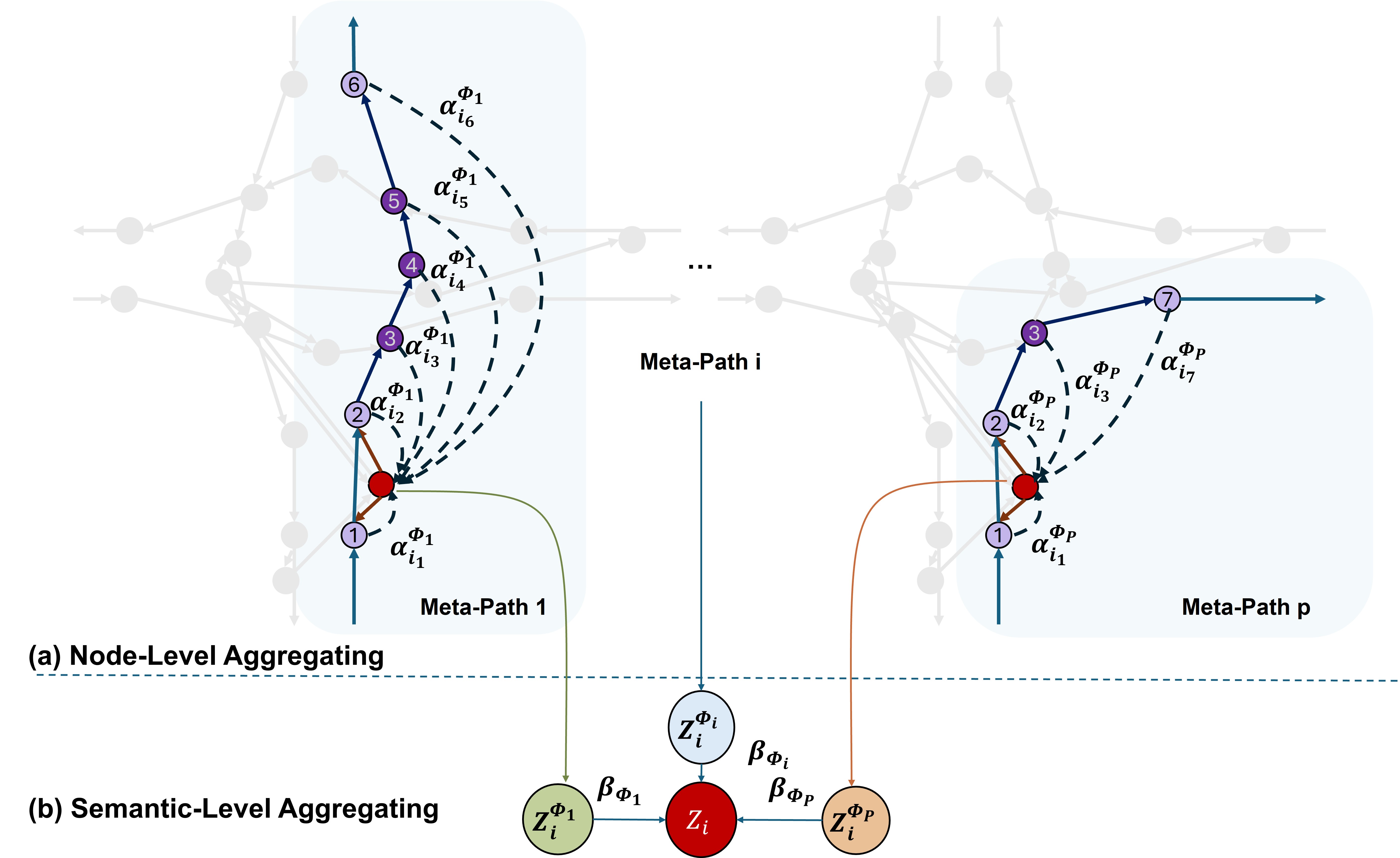} 
  \caption{Illustration of node and semantic levels of attention from the respective of the traffic graph. All traffic participants will receive guidance from corresponding meta-paths.}
    \vspace{-15pt}
  \label{fig:HAN_traffic}
\end{figure}

A heterogeneous graph operator is used to reason over the given scene graph $g_i$. To better incorporate the generated meta-paths, we follow the principle from HAN~\cite{wang2019heterogeneous} i.e. using a hierarchical attention structure from node-level attention to semantic-level attention as shown in figure~\ref{fig:HAN_traffic}.

Applying HAN to learn relational information is shown in Algorithm~\ref{alg:example}. 
Three distinct node types are used for the probability predictor to encode participants, lane snippets, and lane connectors. 
We use $p_i$, $s_j$, $c_z$ to represent these three types respectively, where $p_i \in Z$, $s_j \in Z$, $c_z \in Z$.
\vspace{-10pt}
\begin{algorithm} \small
    \SetAlgoLined
    \SetKwData{Left}{left}\SetKwData{This}{this}\SetKwData{Up}{up} \SetKwFunction{Union}{Union}\SetKwFunction{FindCompress}{FindCompress} \SetKwInOut{Input}{input}\SetKwInOut{Output}{output} \SetKwInOut{Return}{return}
    \Input {Heterogeneous scene graph $G=(V, E, \tau, \phi)$ \\
          Node feature $\left\{h_i, \forall i \in V, h \in \{ p, s, c\}\right\}$ \\
           Meta-path set $\left\{\Phi_0, \Phi_1, \ldots, \Phi_P\right\}$ \\
           Number of attention head $K$
           } 
    \Output {Heterogeneous graph node embedding $Z$}
    \BlankLine 
    \For{$\Phi_i \in\left\{\Phi_0, \Phi_1, \ldots, \Phi_P\right\}$}{
        \For{ $k=1 \ldots K$        
        } {Type-specific transformation $\mathrm{h}_i^{\prime} \leftarrow \text{MLP}\{\mathrm{h}_i$\}

        \For{$i \in V$}{Find the meta-path based neighbors $N_i^{\Phi}$

        \For{$j \in N_i^{\Phi}$}{Calculate the weight coefficient $\alpha_{i j}^{\Phi}$}
        
        Calculate the semantic-specific node embedding $\mathrm{z}_i^{\Phi} \leftarrow \sigma\left(\sum_{j \in N_i^{\Phi}} \alpha_{i j}^{\Phi} \cdot \mathbf{h}_j^{\prime}\right)$
        }
        Concatenate the learned embeddings from all attention head $\mathrm{z}_i^{\Phi} \leftarrow \|_{k=1}^K \sigma\left(\sum_{j \in N_i^{\Phi}} \alpha_{i j}^{\Phi} \cdot \mathbf{h}_j^{\prime}\right)$     
        }
       Calculate the weight of meta-path $\beta_{\Phi_i}$
       Fuse the semantic-specific embedding
       $Z \leftarrow \sum_{i=1}^P \beta_{\Phi_i} \cdot Z_{\Phi_i}$
    }
    \KwRet{$Z$}
    \caption{ \small Semantic Graph Learning via HAN}
       \label{alg:example}
\end{algorithm}
\vspace{-10pt}
\subsubsection{Probability Predictor}
As a result of the scene graph encoder, nodes of lane snippets $s_i$ and lane connectors $c_i$ are projected to the same dimension $Z$. 
We treat these two types of nodes as the same type and use $l_i$ to represent them. Inspired by LAFormer~\cite{Liu2023LAformerTP}, we align the target agent motion and lane information at each future time step $t \in \{1, \ldots, t_f\}$. 
To achieve this, we use a lane score head and an attention mechanism to predict lane encoding probabilities. 
In the attention mechanism, key ($K$) and value ($V$) vectors are produced by $MLP(p_i)$, whereas the query ($Q$) is produced by $MLP(l_i)$. 
Next, attention encodings are calculated by $A_{i, j}=\operatorname{softmax}\left(\frac{Q K^T}{\sqrt{d_k}}\right) V$. 
The predicted score of the $j\text{th}$ lane encodings at $t$ is shown in equation \ref{dense_lane}, where $\phi$ denotes MLP layers. 
We select top-k lane encodings to maintain the uncertainty and concatenate the candidate lane segments and associated scores over the future time steps to obtain $ L = \operatorname{ConCat}\left\{l_{1: k}, \hat{s}_{1: k}\right\}_{t=1}^{t_f}$.
\begin{equation}
\hat{s}_{j, t}=\frac{\exp \left(\phi\left\{p_i, l_j, A_{i, j}\right\}\right)}{\sum_{n=1}^{N_{\text {lane} \in \Phi_j }} \exp \left(\phi\left\{h_i, l_n, A_{i, n}\right\}\right)},
\label{dense_lane}
\end{equation}
To optimize the probability estimation, we use a binary cross-entropy loss $\mathcal{L}_{\text {lane }}$, as shown in equation~\ref{lane_loss}. 
Ground truth lane segment $s_t$ relies on the \textit{isOn} relationship in the knowledge graph. 
Next, a cross-attention operation is performed to further fuse agent and lane information. 
Key and value vectors are $L$, query vector is $p_i$. The updated lane output is $l_{i, \mathrm{att}}$.
\begin{equation}
\mathcal{L}_{\text {lane }}=\sum_{t=1}^{t_{\mathrm{f}}} \mathcal{L}_{\mathrm{CE}}\left(s_t, \hat{s}_t\right)
\label{lane_loss}
\end{equation}
Then we employ a predictor for generating multimodal trajectories. 
This is realized by sampling a latent vector $z$ from a multivariate normal distribution and adding it to the fusion encodings. 
Next, a Laplacian mixture density network (MDN) decoder is used to output a set of trajectories $\sum_{m=1}^M \hat{\pi}_m \operatorname{Laplace}(\mu, b)$. $\hat{\pi}_m$ denotes the probability of each mode and $\sum_{m=1}^M \hat{\pi}_m=1$. $\mu$ and $b$ represent the location and scale parameters of each Laplace component. 
We use an MLP to predict $\hat{\pi}_m$, a GRU to recover the time dimension $t_{\mathrm{f}}$ of the predictions, and two MLPs to predict $\mu$ and $b$. 
The predictor is trained by minimizing a regression loss and a classification loss. Regression loss is computed using the Winner-Takes-All strategy as shown in equation \ref{reg_loss}.
\begin{equation}
\mathcal{L}_{\mathrm{reg}}=\frac{1}{t_f} \sum_{t=1}^{t_f}-\log P\left(Y_t \mid \mu_t^{m^*}, b_t^{m^*}\right)
\label{reg_loss}
\end{equation}
where $Y$ is the ground truth position and $m^*$ represents the best mode which has minimum $L_2$ error among the $M$ predictions. Cross-entropy loss is used to optimize the mode classification as shown in equation \ref{classification_loss}.
\begin{equation}
\mathcal{L}_{\mathrm{cls}}=\sum_{m=1}^M-\pi_m \log \left(\hat{\pi}_m\right) .
\label{classification_loss}
\end{equation}
Several metrics are used to evaluate the deviation from the ground truth, like velocity loss and angle loss, and investigate the influence of different measurements on the predictions.
For the velocity loss, we calculate the ground truth velocity traces $V_t = \|Y_t - Y_{t-1}\|_2$ and prediction velocity traces $\hat{V_t} = \|\mu_t - \mu_{t-1}\|_2$, then velocity loss is shown in equation \ref{velocity_loss}.
\begin{equation}
\mathcal{L}_{\mathrm{velocity}}=\frac{1}{t_f} \sum_{t=1}^{t_f}-\log P\left(V_t \mid \hat{V_t}^{m^*}, b_t^{m^*}\right)
\label{velocity_loss}
\end{equation}
For the angle loss, $X_0$ is used to denote the initial position and we calculate ground truth angle $\theta_t=\arctan 2\left(Y_t-X_0\right)$ and prediction angle $\hat{\theta}_t=\arctan 2\left(\mu_t-X_0\right)$. 
The following equation~\ref{angle_loss} shows the calculation of the loss:
\begin{equation}
\mathcal{L}_{\text {angle }}=\frac{1}{t_f} \sum_{t=1}^{t_f}-\cos \left(\hat{\theta}_t-\theta_t\right)
\label{angle_loss}
\end{equation}
The total loss for the motion prediction is given by \ref{eq:loss_overall}.
\begin{equation}
\mathcal{L}=\lambda_1 \mathcal{L}_{\text {lane }} +  \lambda_2 \mathcal{L}_{\text {velocity}} + \lambda_3 \mathcal{L}_{\text {angle}} + \mathcal{L}_{\text {reg}}+ \mathcal{L}_{\mathrm{cls}}  
\label{eq:loss_overall}
\end{equation}
\subsection{Prediction Refinement}
\vspace{-5pt}
To filter out the unreasonable predictions, we analyze the predicted trajectories by anchor paths~\cite{naumann2023lanelet2}. 
Anchor paths provide possible and permitted trajectories for an agent at a given position in the road network. 
Anchor paths are used to filter out trajectory candidates far from these anchor paths. 
Next, we cluster the remaining trajectory candidates w.r.t. their speed profiles and keep the top candidates closest to the cluster centers. 
For an unfair comparison, we also perform experiments using the ground truth speed profile to get an idea about the relevance of the speed component in the prediction results. 
Details are shown in Algorithm~\ref{alg:example_rule}. 
\vspace{-10pt}
\begin{algorithm} \small
    \SetAlgoLined
    \SetKwData{Left}{left}\SetKwData{This}{this}\SetKwData{Up}{up} \SetKwFunction{Union}{Union}\SetKwFunction{FindCompress}{FindCompress} \SetKwInOut{Input}{input}\SetKwInOut{Output}{output} \SetKwInOut{Return}{return}

    \Input {Predictions $\left\{\mu_{1: t_f}^1, \mu_{1: t_f}^2, \ldots, \mu_{1: t_f}^k\right\}$  \\
            Predicted Probabilities $\{\pi_1, \pi_2, \ldots, \pi_k\}$ \\
            Anchor Paths $\{P_1, P_2, \ldots P_5\}$
           } 
    \Output {Filtered Predictions$\left\{\hat{Y}_{1: t_f}^1, \hat{Y}_{1: t_f}^2, \ldots, \hat{Y}_{1: t_f}^5\right\} $ }
    \BlankLine 
    \If{Ground Truth speed profile $s_{gt}$ available}
       {Calculate the speed profiles $s_{1}$, $s_{2}$, \ldots, $s_{k}$ 
       
       Calculate similarity to $s_{gt} $ using Dynamic Time Warping (DTW)

        Select 5 most similar predictions $\left\{\hat{Y}_{1: t_f}^1, \hat{Y}_{1: t_f}^2, \ldots, \hat{Y}_{1: t_f}^5\right\}$}
    \Else
   {
    \For{$ P_i \in \{P_1, P_2, \ldots P_5\}$}
        {
        \For {$\mu_{1: t_f}^j \in \left\{\mu_{1: t_f}^1, \mu_{1: t_f}^2, \ldots, \mu_{1: t_f}^k\right\}$}
            {   
                Calculate the distance $d_{ij}$ between $P_i$ and $\mu_{1: t_f}^j$        
            }
        For each $i$, select the $min_5 d_{ij}$ and calculate the speed profiles $s_{i1}$, $s_{i2}$, $s_{i3}$, $s_{i4}$, $s_{i5}$.

        Cluster speed profiles $s_{ij}$ using K-means and output the prediction $\hat{Y}_{1: t_f}^i$ closest to the cluster centers.
        }} 
    \KwRet{$\left\{\hat{Y}_{1: t_f}^1, \hat{Y}_{1: t_f}^2, \ldots, \hat{Y}_{1: t_f}^5\right\}\subseteq \left\{\mu_{1: t_f}^1, \mu_{1: t_f}^2, \ldots, \mu_{1: t_f}^k\right\}$}
    \caption{ \small Prediction Refinement}
       \label{alg:example_rule}
\end{algorithm}
\vspace{-15pt}


\section{EXPERIMENTS}
\vspace{-2.5pt}
\subsection{Dataset \& nuScenes Knowledge Graph}
The nuScenes dataset~\cite{Caesar2020nuscenes} is a popular dataset for self-driving cars that is gathered in Boston and Singapore. 
It encompasses 1000 scenes, each lasting 20 seconds, and includes meticulously annotated ground truth details along with high-definition (HD) maps. 
The vehicles within this dataset have 3D bounding boxes manually annotated and published at a rate of 2 Hz. 
For the prediction task, the objective involves leveraging the preceding 2 seconds of object history and the map data to forecast the subsequent 6 seconds. 
We adhere to the standard split provided by the nuScenes benchmark description. 
Using our proposed ontology to the nuScenes dataset, we generate the nuScenes Knowledge Graph including agent and map information as described in \cite{iccvw_leon}. 
Features are provided by the upstream perception components
and the HD map from the nuScenes dataset. Table~\ref{tab:node_attribute} and ~\ref{tab:relation_attribute} list the used feature sets for each node type and each relation type. All features that express a category type are one-hot encoded. 
\vspace{-7.5pt}
\begin{table}[htp!]
    \centering
    \caption{Node Type Features}
    \label{tab:node_attribute}
    \begin{tabular}{l l l}
        \hline
        \textbf{View} & \textbf{Node type} & \textbf{Features} \\
        \hline
        \multirow{2}{*}{Agent} & SceneParticipant & \makecell[l]{Orientation, State, Position, \\ Velocity, Acceleration, \\ Heading Change, \\ Distance to Centerline}\\
                                & Participant &  Type, Size\\
                                & Sequence &  Timestamp\\
                                & Scene &  - \\
        \hline
        \multirow{13}{*}{Map} & LaneSnippet & Length \\
                             & LaneSlice & Width, Center Pose \\
                             & LaneConnector & - \\
                             & OrderedPose & Center Pose \\
                             & Lane & - \\
                             & CarparkArea & - \\
                             & Walkway & - \\
                             & Intersection & - \\
                             & PedCrossingStopArea & - \\
                             & StopSignArea & - \\
                             & TrafficLightStopArea & - \\
                             & TurnStopArea & - \\
                             & YieldStopAre & - \\
        \hline
    \end{tabular}
\end{table}
\vspace{-7.5pt}
\vspace{-7.5pt}
\begin{table}[htp!]
    \centering
    \caption{Relation Type Features}
    \label{tab:relation_attribute}
    \begin{tabular}{l l l}
        \hline
        \textbf{View} & \textbf{Relation type} & \textbf{Features} \\
        \hline
        \multirow{5}{*}{Agent} & hasSceneParticipant &  -\\
                               & inNextScene &  Time Elapsed\\
                               & hasNextScene &  Time Elapsed\\
                               & hasPreviousScene &  Time Elapsed\\
                               & isSceneParticipant & - \\
                                
        \hline
        \multirow{9}{*}{Map} & switchViaDoubleDashedWhite & - \\
                             & switchViaRoadDivider & - \\
                             & switchViaSingleZigzagWhite & - \\
                             & switchViaDoubleSolidWhite & - \\
                             & switchViaSingleSolidYellow & - \\
                             & switchViaSingleSolidWhite & - \\
                             & isSlice/PoseOnStopArea & - \\
                             & connectsIncoming/Outgoing & - \\
                             & hasNextLane/Snippet/Slice & - \\
                            
        \hline
        \multirow{5}{*}{Interaction} & isOnMapElement & Probability \\
                   & relatedLongitudinal &   Path/Distance \\
                   & relatedLateral &  Path/Distance \\
                   & relatedIntersecting & Path/Distance \\
                   & relatedPedestrian & Distance \\
        \hline
    \end{tabular}
\end{table}
\vspace{-7.5pt}
\subsection{Metrics}
We utilize standard evaluation metrics to assess the prediction performance, specifically employing $ADE_K$ (Average Displacement Error for $K$ modes) and $FDE_K$ (Final Displacement Error for $K$ modes). 
These metrics gauge $L_2$ errors, both at the final step and averaged across each step for predicting $K$ modes. 
The reported minimum error among the $K$ modes is considered. 
Both ADE and FDE are measured in meters. 
Additionally, the miss rate $MR_K$ calculates the percentage of scenarios where the final-step error exceeds 2 meters.
\vspace{-5.5pt}
\subsection{Model Implementation}
\vspace{-2pt}
The hidden dimension of vectors in the pipeline is set to 32. 
The layer of the heterogeneous graph neural network is set to 1 and \textit{sum} is used as the aggregation method. 
The attention head in HAN is set to 8 whereas values for parameters of equation \ref{eq:loss_overall}, $\lambda_1$, $\lambda_2$ and $\lambda_3$ are set to 0.95, 1, and 1, respectively.

We use all agent and map elements within the four closest \textit{roadblocks}. 
The coordinate system in the model is the BEV centered at the agent location at $t = 0$. We use the orientation from the agent location at $t = -1$ to the agent location at $t = 0$ as the positive x-axis. 
The model is trained on a TESLA-V100 GPU, with a batch size of 32, and the Adam optimizer with an initial learning rate of $1 \times 10^{-3}$ decayed by 0.7 per 5 epochs. 
\vspace{-5.5pt}
\subsection{Quantitative Results}
\vspace{-2pt}
We compare our results on the nuScenes online benchmark as shown in Table~\ref{tab:performance_nuScenes}. 
The SemanticFormer method predicts directly 5 trajectories without prediction refinement, whereas its extension, SemanticFormerR, predicts 25 trajectories and then refines those predictions. As can be observed, SemanticFormerR achieves competitive performance, thus indicating the benefit of leveraging complex and heterogeneous scene information represented in the Knowledge Graph. Also, it suggests that the speed profiles have a huge impact on future trajectories. 
In an unfair comparison, utilizing ground truth speed followed by Algorithm~\ref{alg:example_rule}, SemanticFormerR demonstrates a significant superiority over state-of-the-art methods.
\vspace{-7.5pt}
\begin{table}[htp!]
    \centering
    \caption{Performance Table on nuScenes Benchmark}
    \label{tab:performance_nuScenes}
    \begin{tabular}{l c c c c c c}
        \hline
        \multirow{2}{*}{\textbf{Method}}  & \multirow{2}{*}{\makecell[c]{\textbf{GT} \\ \textbf{Speed}}} & \multirow{2}{*}{\makecell[c]{\textbf{K=1} \\ \textbf{FDE}}} &  \multicolumn{2}{c}{\textbf{K=5}} &  \multicolumn{2}{c}{\textbf{K=10}} \\
                                &       &  & \textbf{ADE} & \textbf{MR} & \textbf{ADE} & \textbf{MR}\\
        
        \hline
        CoverNet \cite{PhanMinh2019CoverNetMB} & $\times$ & 11.36 &  1.96 &  0.67  & 1.48 & - \\
        Trajectron++ \cite{salzmann2020trajectron++}& $\times$ & 9.52  &  1.88 &  0.70  & 1.51 & 0.57 \\
        LaPred \cite{kim2021lapred} & $\times$ & 8.37 & 1.47 &  0.53  & 1.12 & 0.46 \\
        P2T \cite{Deo2020TrajectoryFI}& $\times$ & 	10.50 & 1.45 &  0.64  & 1.16 & 0.46 \\
        LaneGCN \cite{Liang2020Learning} & $\times$ & - &  - &  0.49  & 0.95 & 0.36 \\
        GOHOME \cite{Gilles2021GOHOMEGH} & $\times$ & 6.99 & 1.42 &  0.57  & 1.15 & 0.47 \\
        Autobot \cite{Girgis2021LatentVS} & $\times$ & 8.19 & 1.37 &  0.62  & 1.03 & 0.44 \\
        THOMAS \cite{Gilles2021THOMASTH} & $\times$ & 6.71 & 1.33 &  0.55  & 1.04 & - \\

        PGP \cite{Deo2021MultimodalTP} & $\times$ & 7.17 & 1.30 &  0.61  & 1.00 & 0.37 \\
        LaFormer \cite{Liu2023LAformerTP} & $\times$ & 6.95 &  1.19 &  0.48  & 0.93 & 0.33 \\
        Socialea \cite{chen2023q}  & $\times$ & 6.77 & 1.18 & 0.48  & 1.02 & 0.44 \\
        FRM \cite{Park2023LeveragingFR}  & $\times$ & 6.59 & 1.18 & 0.48 & 0.88 & 0.30 \\
        
        SemanticFormer & $\times$ & 6.29 & 1.15 &  \textbf{0.48} & 0.91 & 0.31  \\
        SemanticFormerR & $\times$ & \textbf{6.27} & \textbf{1.14} & 0.50 & \textbf{0.87} & \textbf{0.30} \\
        
        \hline
        DMAP \cite{naumann2023lanelet2} & $\checkmark$ & - & 1.09 & \textbf{0.19} & 1.07 & 0.18 \\
                
        SemanticFormerR & $\checkmark$ & 	\textbf{3.88} & \textbf{0.86}  & 0.26  & \textbf{0.78} & \textbf{0.13}  \\
        
        \hline
    \end{tabular}
\end{table}
\vspace{-25pt}
\subsection{Ablation study}
\vspace{-2.5pt}
\subsubsection{Effect of Topological Structure of Heterogeneous Graph}
Knowledge graph provides explicit and logical relationships between different heterogeneous nodes. 
We study the performance improvement compared to fully connected or unconnected graph structure as shown in Table~\ref{tab:ablation_study_graph_structure}.  
\vspace{-7.5pt}
\begin{table}[htp!]
    \centering
    \caption{Ablation Study of Graph Topological Structure}
    \label{tab:ablation_study_graph_structure}
    \begin{tabular}{c c c c }
        \hline
        \multirow{2}{*}{\textbf{Graph Topology}}  & \multirow{2}{*}{\textbf{Edge Types}}  & \multicolumn{2}{c}{\textbf{K=5}} \\
                     & & \textbf{ADE} & \textbf{FDE} \\
        
        \hline
        Knowledge Graph        &  46  & \textbf{1.15} & \textbf{2.20} \\
        Fully Connected Graph  & 1 & 1.19 & 2.31  \\
        Fully Unconnected Graph   & 0 & 1.24 & 2.46   \\

        \hline
    \end{tabular}
\end{table}
\vspace{-5pt}
\subsubsection{Effect of Individual Components}
Our proposed heterogeneous graph is mainly composed of four parts which are map topology, meta-paths, agent-map relationships, and agent-agent relationships. 
We investigate the impact of dropping certain inputs to the model as shown in Table~\ref{tab:ablation_study_graph_component}. 
\vspace{-5pt}
\begin{table}[htp!]
    \centering
    \caption{Ablation Study for Graph Components}
    \label{tab:ablation_study_graph_component}
    \begin{tabular}{c c c c c c}
        \hline
        \multirow{2}{*}{\makecell[c]{\textbf{Meta-} \\ \textbf{Paths}}}     &  
        \multirow{2}{*}{\makecell[c]{\textbf{Map-} \\ \textbf{Topology}}}   &  
        \multirow{2}{*}{\makecell[c]{\textbf{Agent-} \\ \textbf{Map}}}      &
        \multirow{2}{*}{\makecell[c]{\textbf{Agent-} \\ \textbf{Agent}}}    &
        \multicolumn{2}{c}{\textbf{K=5}} \\
            & & & & \textbf{ADE} & \textbf{FDE} \\
            
        \hline
        
           $\checkmark$ & $\checkmark$ & $\checkmark$ & $\checkmark$ & \textbf{1.15} &  \textbf{2.20}  \\
           $\times$ & $\checkmark$ &  $\checkmark$ & $\checkmark$   & 1.18 & 2.29     \\
           $\checkmark$ & $\times$ &  $\checkmark$ & $\checkmark$   & 1.17 &  2.26  \\
           $\times$ & $\times$ & $\checkmark $ & $\checkmark$  & 1.22 & 2.39     \\
           $\times$ & $\times$ & $\times $ & $\checkmark$   & 1.23 & 2.42     \\
           $\times$ & $\times$ & $\times $ & $\times$   &  1.24 & 2.46     \\

        \hline
    \end{tabular}
\end{table}
\vspace{-5pt}
\subsubsection{Integration to other Models}
We integrate our proposed Knowledge Graph into other graph-based models like VectorNet and LaFormer. 
Table~\ref{tab:ablation_study_plugging} shows the experimental results indicating that the Knowledge Graph can effectively improve the performance of the chosen methods.

\vspace{-5pt}
\begin{table}[htp!]
    \centering
    \caption{Ablation Study for Integrating other Architectures}
    \label{tab:ablation_study_plugging}
    \begin{tabular}{c c c c }
        \hline
         Architectures  &  \textbf{ADE\_5} & \textbf{FDE\_1}  & \textbf{OffRoadRate}  \\
        \hline
        VectorNet~\cite{Gao2020VectorNetEH}      &  1.34  & 7.98 & 0.04  \\
        VectorNet + KG  &  \textbf{1.26}  & \textbf{7.55} & \textbf{0.03}     \\
        LaFormer~\cite{Liu2023LAformerTP}      &  1.19 &  6.95  & 0.02   \\
        LaFormer + KG   & \textbf{1.15}  & \textbf{6.29} &  \textbf{0.02}    \\

        \hline
    \end{tabular}
\end{table}
\vspace{-5pt}
\subsubsection{Effect of Heterogeneous Graph Operators}
We analyze different heterogeneous graph operators like HGT~\cite{hu2020heterogeneous} and HAN~\cite{wang2019heterogeneous}. 
As shown in Table~\ref{tab:ablation_study_1}, to prevent overfitting, we merge sub-classes like single solid, double solid, etc, to \textit{switchViaPermitted} and \textit{switchViaNonPermitted} relationships. *N means number of layers of operator is N.
\begin{table}[htp!]
    \centering
    \caption{Ablation Study for HGNN Operators}
    \label{tab:ablation_study_1}
    \begin{tabular}{c c c c c c}
        \hline
        \multirow{2}{*}{\makecell[c]{\textbf{Interaction}\\ \textbf{Graph}}}  &  \multirow{2}{*}{\makecell[c]{\textbf{Oper-}\\ \textbf{ators}}}  & \multirow{2}{*}{\makecell[c]{\textbf{Self}\\\textbf{Loop}}}  & \multirow{2}{*}{\makecell[c]{\textbf{Meta}\\\textbf{Path}}}  &  \multicolumn{2}{c}{\textbf{K=5}} \\
                                       & & &  &  \textbf{ADE} & \textbf{FDE} \\
        
        \hline
        Original & $\mathrm{HGT}^* 2$ &  $\checkmark$ & $\times$ &  1.24 &  2.49  \\
        Compact & $\mathrm{HGT}^* 2$ &  $\checkmark$ & $\times$ &  1.24 &  2.46  \\
        Compact & $\mathrm{HGT}^* 2$ &  $\times$ & $\times$ &   1.22 &  2.38  \\
        Compact & $\mathrm{HAN}^* 2$ &  $\times$ & $\checkmark$ & 1.19 &  2.34  \\
        Compact & $\mathrm{HAN}^* 1$ &  $\times$ & $\checkmark$ & 1.15 &  2.20  \\
        \hline
    \end{tabular}
\end{table}
\vspace{-5pt}
\subsection{Qualitative results}
\begin{figure}[htp!]
    \centering
    \includegraphics[width=0.45\textwidth]{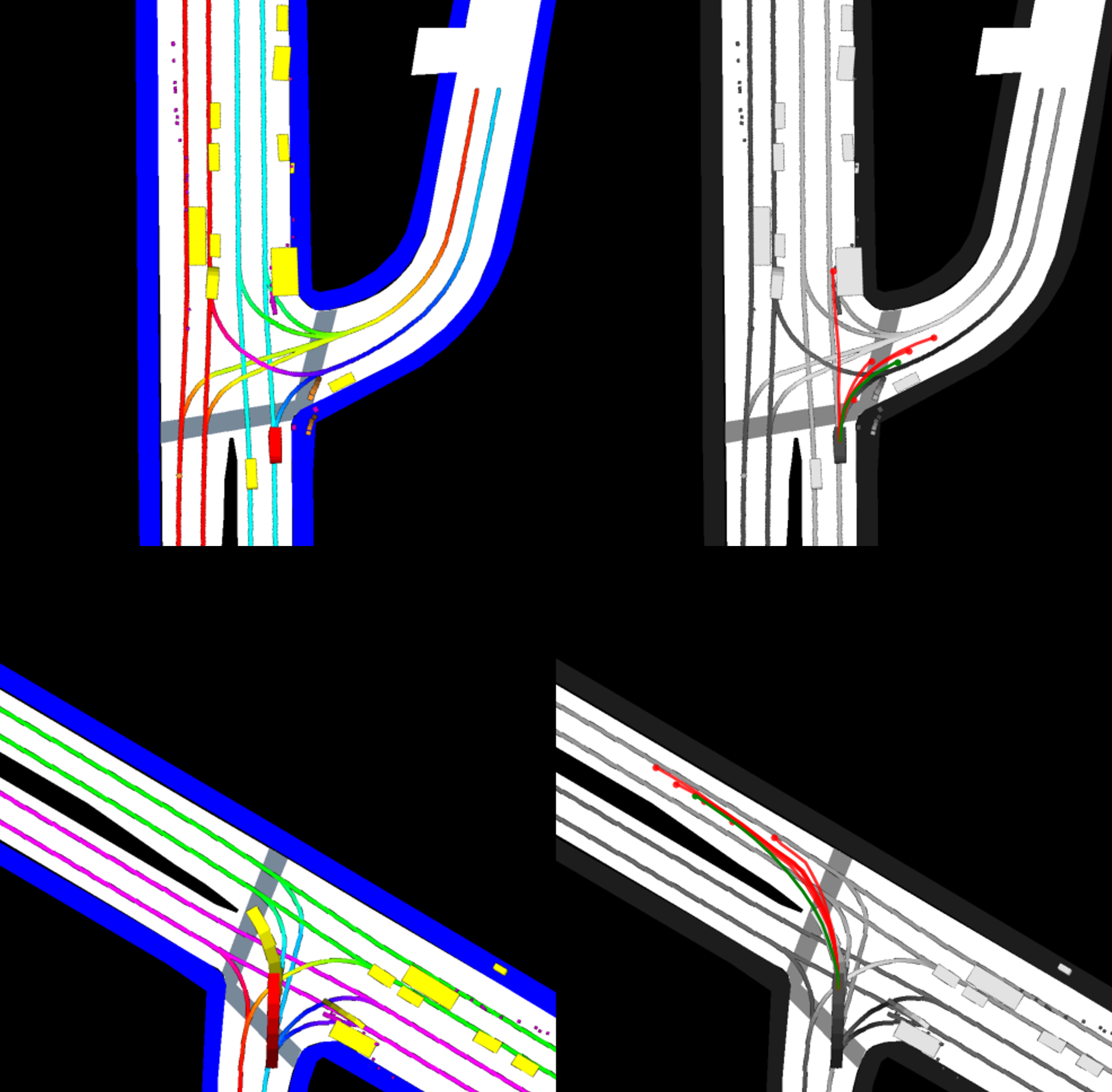} 
    \caption{Illustration of the qualitative result.
    Column 1 is the traffic scene and column 2 is the results of SemanticFormerR.}
    \label{fig:qualitative_results}
\end{figure}

A qualitative visualization of our predictions is depicted in Figure~\ref{fig:qualitative_results}.
Green trajectories are ground truth and red trajectories are five predictions. 
Row 1 shows predictions considering all driving path possibilities and row 2 captures the lane-changing situation successfully.

\section{CONCLUSIONS}
This paper proposes a novel approach using a traffic scene knowledge graph leveraging past trajectories and an HD map as input for predicting a set of multimodal trajectories.
A scene graph encoder module aims to capture the interactions in a traffic scene from four aspects, agent-agent interaction, agent-map interaction, map-map interaction, and meta-paths interaction. 
Further, the refinement module considers the typical speed profiles and anchor paths to refine trajectory candidates.
Our approach achieves excellent results compared to the state-of-the-art model, We also provide an experimental justification of our approach by performing experiments with two SOTA methods, i.e. LaFormer and VectorNet, and replacing their original homogeneous graphs with our Knowledge Graph. We show that the Knowledge Graph improves the performance of those methods by 5\% and 4\%, respectively. 
Moreover, extensive ablation and sensitivity studies also indicate that our proposed Knowledge Graph can be easily integrated into other graph-based methods to improve performance. Future work will focus on extending the Knowledge Graph with additional information such as traffic rules, traffic signs, and forms of driving common sense knowledge.





\bibliographystyle{IEEEtran}
\bibliography{IEEEabrv,IEEEfull.bib,biblio.bib}

\end{document}